\documentclass[11pt]{article}
\usepackage{eacl2014}
\usepackage{times}
\usepackage{url}
\usepackage{latexsym}
\usepackage[usenames,dvipsnames]{color}

\title{Simple, Robust and (almost) Unsupervised Generation of Polarity Lexicons for
  Multiple Languages}

\author{I\~naki San Vicente, Rodrigo Agerri, German Rigau\\
IXA NLP Group\\
University of the Basque Country (UPV/EHU)\\
Donostia-San Sebasti\'an\\
{\{\tt inaki.sanvicente,rodrigo.agerri,german.rigau\}@ehu.eus}
}
\date{}

\begin{document}
\maketitle
\begin{abstract}
This paper presents a simple, robust and (almost) unsupervised dictionary-based method,
\emph{qwn-ppv} (Q-WordNet as Personalized PageRanking Vector) to
automatically generate polarity lexicons. We show that
\emph{qwn-ppv} outperforms other automatically generated lexicons for
the four extrinsic evaluations presented here. It
also shows very competitive and robust results
with respect to manually annotated ones. Results suggest that no single
lexicon is best for every task and dataset and that the intrinsic
evaluation of polarity lexicons is not a good performance indicator on a Sentiment
Analysis task. The \emph{qwn-ppv} method allows to easily create
quality polarity lexicons whenever no domain-based annotated corpora
are available for a given language.
\end{abstract}

\section{Introduction}\label{sec:intro}

Opinion Mining and Sentiment Analysis are important for determining
opinions about commercial products, on companies reputation
management, brand monitoring, or to track attitudes by mining social
media, etc. Given the explosion of information produced and
shared via the Internet, it is not possible to keep up with the
constant flow of new information by manual methods.

Sentiment Analysis often relies on the availability of words and
phrases annotated according to the positive or negative connotations
they convey. `Beautiful', `wonderful', and `amazing' are examples of
positive words whereas `bad', `awful', and `poor' are examples of
negatives.


The creation of lists of sentiment words has generally been performed
by means of manual-, dictionary- and corpus-based methods. Manually
collecting such lists of polarity annotated words is labor intensive and
time consuming, and is thus usually combined with automated approaches
as the final check to correct mistakes. However, there are well known
lexicons which have been fully
\cite{Stoneetal:1966,taboada_lexicon-based_2010} or at least partially
\emph{manually created} \cite{hu_mining_2004,RiloffWiebe:2003}.

\emph{Dictionary-based} methods rely on some dictionary or lexical knowledge
base (LKB) such as WordNet \cite{Wordnet:1998} that contain synonyms
and antonyms for each word. A simple technique in this approach is to
start with some sentiment words as seeds which
are then used to perform some iterative propagation on the
LKB \cite{hu_mining_2004,KimHovy:2004,takamura_extracting_2005,turney_measuring_2003,mohammad_generating_2009,agerri_q-wordnet:_2010,baccianella_sentiwordnet_2010}.

\emph{Corpus-based} methods have usually been applied to obtain
domain-specific polarity lexicons: they have been created by either starting from a seed list of
known words and trying to find other related words in a corpus or
by attempting to directly adapt a given lexicon to a new one using a
domain-specific corpus
\cite{hatzivassiloglou_predicting_1997,turney_measuring_2003,ding_holistic_2008,choi_adapting_2009,mihalcea_learning_2007}. One
particular issue arising from corpus methods is that
for a given domain the same word can be positive in one context but
negative in another. This is also a problem shared by manual
and dictionary-based methods, and that is why \emph{qwn-ppv} also
produces synset-based lexicons for approaches on Sentiment Analysis at
sense level.



This paper presents a simple, robust and (almost) unsupervised dictionary-based method,
\emph{QWordNet-PPV} (QWordNet by Personalized PageRank Vector) to
automatically generate polarity lexicons based on propagating some
automatically created seeds using a Personalized PageRank algorithm
\cite{agirre2014random,AgirreSoroa:2009} over a LKB projected into a graph. We see
\emph{qwn-ppv} as an effective methodology to easily 
create polarity lexicons for any language for which a WordNet is
available.   

This paper empirically shows that: (i) \emph{qwn-ppv} outperforms
other automatically generated lexicons (e.g. SentiWordNet 3.0, MSOL) on 
the 4 extrinsic evaluations presented here; it also displays competitive and robust
results also with respect to manually annotated lexicons; (ii) no
single polarity lexicon is fit for every Sentiment Analysis
task; depending on the text data and the task itself, one lexicon will
perform better than others; (iii) if required, 
\emph{qwn-ppv} efficently generates many lexicons on demand, depending
on the task on which they will be used; (iv) intrinsic evaluation is not
appropriate to judge whether a polarity lexicon is fit for a given
Sentiment Analysis (SA) task because 
good correlation with respect to a \emph{gold-standard} does not
correspond with \emph{correlation} with respect to a SA task;   (v) it is easily applicable
to create \emph{qwn-ppv(s)} for \emph{other languages}, and we
demonstrate it here by creating many polarity lexicons not only for
English but also for Spanish; (vi) the method works at \emph{both word and sense} levels and
it only requires the availability of a LKB or dictionary; finally, (vii) a dictionary-based method like
\emph{qwn-ppv} allows to easily create quality polarity lexicons
whenever no domain-based annotated reviews are available for a given
language. After all, there usually is available a dictionary for a
given language; for example, the Open Multilingual WordNet site lists WordNets for up to 57
languages \cite{bond_linking_2013}.

Although there has been previous work using graph methods for obtaining
lexicons via propagation, the \emph{qwn-ppv} method to combine the seed
generation and the Personalized PageRank propagation is novel.
Furthermore, it is considerable simpler and obtains better and easier to
reproduce results than previous automatic approaches
\cite{EsuliSebastiani:2007,mohammad_generating_2009,rao_semi-supervised_2009}. 

Next section reviews previous related work, taking special interest on those that are currently
available for evaluation purposes. Section
\ref{sec:extr-polar-from} describes the \emph{qwn-ppv} method to
automatically generate lexicons. The resulting lexical
resources are evaluated in section \ref{sec:evaluation}. We finish
with some concluding remarks and future work in section
\ref{sec:concluding-remarks}.

\section{Related Work}\label{sec:previous}

There is a large amount of work on Sentiment Analysis and Opinion
Mining, and good comprehensive overviews are already available
\cite{pang_opinion_2008,liu_sentiment_2012}, so we
will review the most representative and closest to the present work.
This means that we will not be reviewing corpus-based approaches
but rather those constructed manually or upon a dictionary or LKB. We
will in turn use the approaches here reviewed for comparison with
\emph{qwn-ppv} in section \ref{sec:evaluation}.

The most popular manually-built polarity lexicon is part of the General
Inquirer \cite{Stoneetal:1966}, and consists of 1915 words labelled
as ``positive'' and 2291 as ``negative''. Taboada \textit{et al.}
\shortcite{taboada_lexicon-based_2010} manually created their lexicons
annotating the polarity of 6232 words on a scale of 5 to -5. Liu
\textit{et al.}, starting with Hu and Liu \shortcite{hu_mining_2004}, have along
the years collected a manually corrected polarity lexicon
which is formed by 4818 negative and 2041 positive words. Another
manually corrected lexicon \cite{RiloffWiebe:2003} is the one used by the Opinion
Finder system \cite{wilson_recognizing_2005} and contains 4903
negatively and 2718 positively annotated words respectively.

Among the automatically built lexicons, Turney and Littman
\shortcite{turney_measuring_2003} proposed a minimally supervised algorithm to
calculate the polarity of a word depending on whether it
co-ocurred more with a previously collected small set of positive words
rather than with a set of negative ones. Agerri and Garc\'ia Serrano
presented a very simple method to extract the polarity information starting
from the \textit{quality} synset in WordNet
\cite{agerri_q-wordnet:_2010}. Mohammad \textit{et al.} \shortcite{mohammad_generating_2009} 
developed a method in which they 
first identify (by means of affixes rules) a set of positive/negative
words which act as seeds, then used a Roget-like thesaurus to mark
the synonymous words for each polarity type and to generalize from the
seeds. They produce several lexicons the best of which, MSOL(ASL and
GI) contains 51K and 76K entries respectively and uses the full General
Inquirer as seeds. They performed both
intrinsic and extrinsic evaluations using the MPQA 1.1 corpus.

Finally, there are two approaches that are somewhat closer to us, because
they are based on WordNet and graph-based methods. SentiWordNet 3.0 \cite{baccianella_sentiwordnet_2010}
is built in 4 steps: (i) they select the synsets of 14 paradigmatic positive and negative words used as seeds
\cite{turney_measuring_2003}. These seeds are then iteratively extended following the construction of
WordNet-Affect \cite{StrapparavaValitutti:2004}. (ii) They train 7
supervised classifiers with the synsets' glosses which are used to assign
\emph{polarity} and \emph{objectivity} scores to WordNet
senses. (iii) In SentiWordNet 3.0 \cite{EsuliSebastiani:2007}
they take the output of the supervised classifiers as input to applying PageRank
to WordNet 3.0's graph. (iv) They intrinsically evaluate it with respect to
MicroWnOp-3.0 using the \emph{p-normalized Kendall $\tau$ distance}
\cite{baccianella_sentiwordnet_2010}. Rao and Ravichandran \shortcite{rao_semi-supervised_2009} apply different
semi-supervised graph algorithms (Mincuts, Randomized
Mincuts and Label Propagation) to a set of seeds
constructed from the General Inquirer. They evaluate the generated
lexicons intrinsically taking the General Inquirer as the gold
standard for those words that had a match in the generated
lexicons. 

In this paper, we describe two methods to automatically generate seeds
either by following Agerri and Garc\'ia-Serrano \shortcite{agerri_q-wordnet:_2010} or
using Turney and Littman's \shortcite{turney_measuring_2003}
seeds. The automatically obtained seeds are then fed into a
Personalized PageRank algorithm which is applied over a WordNet projected on
a graph. This method is fully automatic, simple and unsupervised
as it only relies on the availability of a LKB.

\section{Generating qwn-ppv}\label{sec:extr-polar-from}

The overall procedure of our approach consists of two steps: \textbf{(1)} 
automatically creates a set of seeds by iterating over a LKB (e.g. a
WordNet) relations; and \textbf{(2)} uses the seeds to initialize
contexts to propagate over the LKB graph using a Personalized Pagerank
algorithm. The result is \emph{qwn-ppv(s)}: Q-WordNets as Personalized PageRanking Vectors.

\subsection{Seed Generation}\label{sec:qwn:-seed-generation}

We generate seeds by means of two different automatic procedures.

\begin{enumerate}
\item \textbf{AG}: We start at the \emph{quality synset} of WordNet and iterate
  over WordNet relations following the original Q-WordNet method described in Agerri and
  Garc\'ia Serrano \shortcite{agerri_q-wordnet:_2010}. 
\item \textbf{TL}: We take a short manually created list of 14 positive and
  negative words \cite{turney_measuring_2003} and iterate over WordNet
  using five relations: \textit{antonymy, similarity, derived-from,
    pertains-to and also-see}.
\end{enumerate}

The \textbf{AG} method starts the propagation from the
attributes of the \emph{quality synset} in WordNet. There
are five noun quality senses in WordNet, two of which contain
attribute relations (to adjectives). From the
\textit{quality$^1_n$} synset the attribute relation takes us to
\textit{positive$^1_a$}, \textit{negative$^1_a$},
\textit{good$^1_a$} and \textit{bad$^1_a$}; \textit{quality$^2_n$}
leads to the attributes \textit{superior$^1_a$} and
\textit{inferior$^2_a$}. The following step is to iterate through
every WordNet relation collecting (i.e., annotating) those synsets that are
accessible from the seeds. Both \emph{AG} and \emph{TL} methods to
generate seeds rely on a number of relations to obtain a more balanced
POS distribution in the output synsets. The output of both methods is a list 
of (assumed to be) positive and
negative synsets. Depending on the number of iterations performed a
different number of seeds to feed UKB is obtained. Seed numbers vary from 100 hundred to 10K
synsets. Both seed creation methods can be applied to any WordNet, not
only Princeton WordNet, as we show in section \ref{sec:evaluation}.

\subsection{PPV generation}\label{sec:ppv-generation}

\begin{table*}[!t]\scriptsize
\centering
\begin{tabular}{ccccccccccccccc} \hline
& \multicolumn{7}{c}{\textbf{Synset Level}} & \multicolumn{7}{c}{\textbf{Word level}}\\ \hline \hline
 & & \multicolumn{3}{c}{Positives} & \multicolumn{3}{c}{Negatives} &  & \multicolumn{3}{c}{Positives} & \multicolumn{3}{c}{Negatives}\\
\textbf{Lexicon} & size & P & R & F & P & R & F & size & P & R & F & P & R & F\\ \hline \hline
\multicolumn{15}{l}{\textit{\scriptsize{Automatically created}}} \\
MSOL(ASL-GI)* & 32706 & .65 & .45 & .53 & .58 & .76 & .66 & 76400 & .70 & .49 & .58 & .61 & .79 & .69 \\
QWN & 15508 & .69 & .53 & .60 & .62 & .76 & .68 & 11693 & .64 & .53 & .58 & .60 & .70 & .65\\
SWN & 27854 & .73 & .57 & .64 & .65 & .79 & .71 & 38346 & .70 & .55 & .62 & .63 & .77 & .69 \\
QWN-PPV-AG(s03\_G1/w01\_G1) & 2589 & .77 & .63 & .69 & .69 & .81 & .74 & 5119 & .68 & .77 & \textbf{.72} & .73 & .64 & \textbf{.68} \\ 
\textbf{QWN-PPV-TL}(s04\_G1/w01\_G1) & 5010 & .76 & .66 & \textbf{.70} & .70 & .79 & \textbf{.74} & 4644 & .68 & .71 & .69 & .70 & .67 & .68\\ \hline\hline
\multicolumn{15}{l}{\scriptsize{\textit{(Semi-) Manually created}}}\\
GI* & 2791 & .74 & .57 & .64 & .65 & .80 & .72 & 3376 & .79 & .64 & .71 & .70 & .83 & .76 \\
OF* & 4640 & .77 & .61 & .68 & .68 & .81 & .74 & 6860 & .82 & .71 & .76 & .74 & .84 & .79 \\
\textbf{Liu*} & 4127 & .81 & .63 & \textbf{.71} & .70 & .85 & \textbf{.76} & 6786 & .85 & .74 & \textbf{.79} & .77 & .87 & \textbf{.82}\\
SO-CAL* & 4212 & .75 & .57 & .64 & .65 & .81 & .72 & 6226 & .82 & .70 & .76 & .74 & .85 & .79 \\ \hline
\end{tabular}
\caption{Evaluation of lexicons at document level using Bespalov's Corpus.}\label{tab:evalbespalov}
\end{table*}

The second and last step to generate \emph{qwn-ppv}(s) consists of
propagating over a WordNet graph to obtain a Personalized PageRanking Vector
(PPV), one for each polarity. This step requires:

\begin{enumerate}
\item A LKB projected over a graph.
\item A Personalized PageRanking algorithm which is applied over the graph.
\item Seeds to create contexts to start the propagation, either words
  or synsets.
\end{enumerate}

Several undirected graphs based on WordNet 3.0 as represented by the MCR 3.0
\cite{agirre2012multilingual} have been created for the
experimentation, which correspond to 4 main sets:  (G1) two
graphs consisting of every synset linked by the
\emph{synonymy} and \emph{antonymy} relations; (G2) a graph with the nodes
linked by every relation, including glosses; (G3) a graph consisting of
the synsets linked by every relation except those that are linked by
\emph{antonymy}; finally, (G4) a graph consisting of the nodes related by every
relation except the \emph{antonymy} and \emph{gloss} relations.

Using the (G1) graphs, we propagate from the seeds over each type
of graph (synonymy and antonymy) to obtain 
two rankings per polarity. The graphs created in (G2), (G3) and (G4) are used to 
obtain two ranks, one for each polarity by propagating from the seeds. In all
four cases the different polarity rankings have to be combined in
order to obtain a final polarity lexicon: the polarity score \textit{pol(s)} of a given
synset \textit{s} is computed by adding its scores in the positive
rankings and subtracting its scores in the negative rankings. If \textit{pol(s)} $> 0$
then \textit{s} is included in the final lexicon as positive. If
\textit{pol(s)} $< 0$ then \textit{s} is included in the final lexicon
as negative. We assume that synsets with \emph{null} polarity scores
have no polarity and consequently they are excluded from the final
lexicon.

The Personalized PageRanking propagation is
performed starting from both synsets and words and using both
\emph{AG} and \emph{TL} styles of seed generation, as explained in 
section \ref{sec:qwn:-seed-generation}. Combining the various
possibilities will produce at least 6 different lexicons for each
iteration, depending on which decisions are taken about which graph, seeds and word/synset to
create the \emph{qwn-ppv}(s). In fact, the experiments produced
hundreds of lexicons, according to the different
iterations for seed generation\footnote{The total time to generate the
final 352 QWN-PPV propagations amounted to around two hours of processing time in a
standard PC.}, but we will only refer to those that obtain the best 
results in the extrinsic evaluations.

With respect to the algorithm to
propagate over the WordNet graph from the automatically created seeds,
we use a Personalized PageRank algorithm \cite{agirre2014random,AgirreSoroa:2009}. 
The famous PageRank \cite{brin_anatomy_1998} algorithm is a
method to produce a rank from the vertices in a graph according to
their relative structural importance. PageRank has also been viewed as
the result of a Random Walk process, where the final rank of a given
node represents the probability of a random
walk over the graph which ends on that same node. Thus, if we
take the created WordNet graph $G$ with $N$ vertices $v_1, \ldots,
v_n$ and $d_i$ as being the outdegree of node $i$, plus a $N \times N$
transition probability matrix $M$ where $M_{ji} = 1/d_i$  if a link
from $i$ to $j$ exists and 0 otherwise, then calculating the PageRank
vector over a graph $G$ amounts to solve the following equation
(\ref{eq:pr}):

\begin{equation}
\mathbf{Pr} = cM\mathbf{Pr} + (1-c)\mathbf{v}
\label{eq:pr}
\end{equation}

In the traditional PageRank, vector \textbf{v} is a uniform normalized vector whose
elements values are all $1/N$, which means that all nodes in the graph
are assigned the same probabilities in case of a random
walk. Personalizing the PageRank algorithm in this case means that it
is possible to make vector \textbf{v} non-uniform and assign stronger
probabilities to certain nodes, which would make the algorithm to
propagate the initial importance of those nodes to their
vicinity. Following Agirre \textit{et al.} \shortcite{agirre2014random}, in
our approach this translates into initializing vector \textbf{v} with those senses
obtained by the seed generation methods described above in section
\ref{sec:qwn:-seed-generation}. Thus, the initialization of vector
\textbf{v} using the seeds allows the Personalized propagation to assign greater
importance to those synsets in the graph identified as being positive
and negative, which resuls in a PPV with the weigths skewed towards
those nodes initialized/personalized as positive and negative. 

\section{Evaluation}\label{sec:evaluation}

\begin{table*}[!t]\scriptsize
\centering
\begin{tabular}{ccccccccccccccc} \hline
& \multicolumn{7}{c}{\textbf{Synset Level}} & \multicolumn{7}{c}{\textbf{Word level}}\\ \hline \hline
& & \multicolumn{3}{c}{Positives} & \multicolumn{3}{c}{Negatives} & & \multicolumn{3}{c}{Positives} & \multicolumn{3}{c}{Negatives}\\
\textbf{Lexicon} & size & P & R & F & P & R & F & size & P & R & F & P & R & F\\ \hline \hline
\multicolumn{15}{l}{\textit{\scriptsize{Automatically created}}} \\
\textbf{MSOL(ASL-GI)}* & 32706 & .56 & .37 & .44 & .76 & .87 & .81 & 76400 & .67 & .5 & .57 & .80 & .89 & .85 \\
QWN & 15508 & .63 & .22 & .33 & .73 & .94 & .83 & 11693 & .58 & .22 & .31 & .73 & .93 & .82 \\
SWN & 27854 & .57 & .33 & .42 & .75 &.89 & .81 & 38346 & .55 & .55 & .55 & .80 & .8 & .80 \\
QWN-PPV-AG (w10\_G3/s09\_G4) & 117485 & .60 & .63 &  {\bf .62} & .83 & .82 &  {\bf .83} & 144883 & .65 & .50 & .57 & .80 & .88 & .84 \\ 
QWN-PPV-TL (s05\_G4) & 114698 & .61 & .58 & .59 & .82 & .83 & .83 & 144883 & .66 & .53 & {\bf  .59} & .81 & .88 & {\bf .84}\\ \hline \hline
\multicolumn{15}{l}{\scriptsize{\textit{(Semi-) Manually created}}}\\
GI* & 2791 & .70 & .32 & .44 & .76 & .94 & .84 & 3376 & .71 & .56 & .62 & .82 & .90 & .86 \\
\textbf{OF}* & 4640 & .67 & .37 & {\bf .48} & .77 & .92 & {\bf .84} & 6860 & .75 & .68 & {\bf .71} & .87 & .90 & {\bf .88} \\
Liu* & 4127 & .67 & .33 & .44 & .76 & .93 & .83 & 6786 & .78 & .45 & .57 & .79 & .94 & .86\\
SO-CAL* & 4212 & .69 & .3 & .42 & .75 & .94 & .84 & 6226 & .73 & .53 & .61 & .81 & .91 & .86 \\ \hline
\end{tabular}
\caption{Evaluation of lexicons using \textit{averaged ratio} on the
  MPQA 1.2$_{test}$ Corpus.}\label{tab:evalmpqa}
\end{table*}

Previous approaches have provided
intrinsic evaluation \cite{mohammad_generating_2009,rao_semi-supervised_2009,baccianella_sentiwordnet_2010}
using manually annotated resources such as the General Inquirer
\cite{Stoneetal:1966} as gold standard. To facilitate
comparison, we also provide such evaluation in section
\ref{sect:intrinsic-eval}. Nevertheless, and as demonstrated by the
results of the extrinsic evaluations, we believe
that polarity lexicons should in general be evaluated \emph{extrinsically}.  After all,
any polarity lexicon is as good as the results obtained by using it
for a particular Sentiment Analysis task.

Our goal is to evaluate the polarity lexicons simplifying the evaluation parameters
to avoid as many external influences as possible on the results. We compare our work 
with most of the lexicons reviewed in
section \ref{sec:previous}, both at synset and word level, both manually and
automatically generated: General Inquirer (\textbf{GI}), Opinion
Finder (\textbf{OF}), \textbf{Liu}, Taboada \textit{et al.'s}
(\textbf{SO-CAL}), Agerri and Garc\'ia-Serrano
\shortcite{agerri_q-wordnet:_2010} (\textbf{QWN}), Mohammad \textit{et
  al's}, (\textbf{MSOL(ASL-GI)}) and
SentiWordNet 3.0 (\textbf{SWN}). The results presented in section
\ref{sec:results} show that \emph{extrinsic} evaluation is more
meaningful to determine the adequacy of a polarity lexicon for a
specific Sentiment Analysis task. 

\subsection{Datasets and Evaluation System}\label{sec:prep-tests-eval}

Three different corpora were
used: Bespalov \textit{et al.'s} \shortcite{bespalov_sentiment_2011} and 
MPQA \cite{RiloffWiebe:2003} for English, and
HOpinion\footnote{http://clic.ub.edu/corpus/hopinion} in Spanish. In addition, we divided the corpus
into two subsets (75\% development and 25\% test) for applying our ratio
system for the phrase polarity task too. Note that the development set
is only used to set up the polarity classification task, and that the
generation of \emph{qwn-ppv} lexicons is unsupervised.   

For Spanish we tried to reproduce the English settings with Bespalov's corpus. Thus,
both development and test sets were created from the HOpinion
corpus. As it contains a much higher
proportion of positive reviews, we created also subsets which contain a
balanced number of positive and negative reviews to allow for a more
meaningful comparison than that of table \ref{tab:evalspanishfull}. Table
\ref{tab:datasets} shows the number of documents per polarity for
Bespalov's, MPQA 1.2 and HOpinion.

\begin{table}[ht]\footnotesize
\centering
\begin{tabular}[ht]{p{2.5cm}ccc}
\hline
Corpus & POS docs & NEG docs & Total\\ \hline \hline
Bespalov$_{dev}$ & 23,112 & 23,112 & 46,227 \\
Bespalov$_{test}$ & 10,557 & 10,557 & 21,115\\
MPQA 1.2$_{dev}$ & 2,315 & 5,260 & 7,575 \\
MPQA 1.2$_{test}$ & 771 & 1,753 & 2,524 \\ 
MPQA 1.2$_{total}$ & 3,086 & 7,013 & 10,099 \\ \hline \hline
HOpinion\_Balanced$_{dev}$ & 1,582 & 1,582 & 3,164 \\
HOpinion\_Balanced$_{test}$ & 528 & 528 & 1,056\\
HOpinion$_{dev}$ & 9,236 & 1,582 & 10,818 \\
HOpinion$_{test}$ & 3,120 & 528 & 3,648\\ \hline
\end{tabular}
\caption{Number of positive and negative documents in train and test sets.}\label{tab:datasets}
\end{table}

We report results of 4 extrinsic evaluations or tasks, three of them based on a simple
\emph{ratio average system}, inspired by Turney
\shortcite{turney_thumbs_2002}, and another one based on Mohammad \textit{et al.}
\shortcite{mohammad_generating_2009}. We first implemented 
a simple \emph{average ratio classifier} which computes the average ratio of the
polarity words found in document $d$:

\begin{equation}
polarity(d) = \frac{\sum_{w \in d} pol(w)}{|d|}
\label{eq1}
\end{equation}

\begin{table*}[!t]\scriptsize
\centering
\begin{tabular}{ccccccccccccccc} \hline
\multicolumn{8}{c}{\textbf{Synset Level}} & \multicolumn{7}{c}{\textbf{Word level}} \\ \hline \hline
 & & \multicolumn{3}{c}{Positives} & \multicolumn{3}{c}{Negatives} & &  \multicolumn{3}{c}{Positives} & \multicolumn{3}{c}{Negatives}\\
\textbf{Lexicon} & size & P & R & F & P & R & F & size & P & R & F & P & R & F\\ \hline \hline
\multicolumn{15}{l}{\textit{\scriptsize{Automatically created}}} \\
MSOL(ASL-GI)* & 32706 & .52 & .48 & .50 & .85 & .62 & .71 & 76400 & .68 & .56 & .62 & .82 & .86 & .84 \\
QWN & 15508 & .50 & .36 & .42 & .84 & .32 & .46 & 11693 & .45 & .49 & .47 & .78 & .51 & .61 \\
SWN & 27854 & .50 & .45 & .47 & .85 & .48 & .61 & 38346 & .49 & .52 & .50 & .78 & .68 & .73 \\
{\bf QWN-PPV-AG} (s09\_G3/w02\_G3) & 117485 & .59 & .67 & {\bf .63} & .85 & .78 & {\bf .82} & 147194 & .64 & .64 & .64 & .84 & .83 & .83 \\ 
\textbf{QWN-PPV-TL} (w02\_G3/s06\_G3) & 117485 & .59 & .57 & .58 & .82 & .81 & .81 & 147194 & .63 & .67 & {\bf .65} & .85 & .81 & \textbf{.83} \\ \hline\hline
\multicolumn{15}{l}{\scriptsize{\textit{(Semi-) Manually created}}}\\
GI* & 2791 & .60 & .40 & .47 & .91 & .38 & .54 & 3376 & .70 & .60 & .65 & .93 & .52 & .67 \\
\textbf{OF}* & 4640 & .63 & .42 & \textbf{.50} & .93 & .46 & \textbf{.62} & 6860 & .75 & .71 & \textbf{.73} & .95 & .66 & \textbf{.78} \\
Liu* & 4127 & .65 & .36 & .47 & .94 & .45 & .60 & 6786 & .78 & .49 & .60 & .97 & .61 & .75 \\
SO-CAL* & 4212 & .65 & .37 & .47 & .92 & .45 & .60 & 6226 & .73 & .57 & .64 & .96 & .59 & .73\\ \hline
\end{tabular}
\caption{Evaluation of lexicons at phrase level using Mohammad
  \textit{et al.'s} \shortcite{mohammad_generating_2009} method on MPQA 1.2$_{total}$ Corpus.}\label{tab:evalmpqaphrase}
\end{table*}

where, for each polarity, \textit{pol(w)} is $1$ if \textit{w} is
included in the polarity lexicon and $0$ otherwise. Documents that reach a certain threshold
are classified as positive, and otherwise as negative. To setup an
evaluation enviroment as fair as possible for every lexicon,
the threshold is optimised by maximising accuracy over the development
data.

Second, we implemented a phrase polarity task identification as
described by Mohammad \textit{et al.}
\shortcite{mohammad_generating_2009}. Their method consists of: (i) if any of the
words in the target phrase is contained in the negative lexicon, then
the polarity is negative; (ii) if none of the words are negative, and
at least one word is in the positive lexicon, then is positive; (iii)
the rest are not tagged.

We chose this very simple polarity estimators because our aim was to minimize the role other
aspects play in the evaluation and focus on how, other things
being equal, polarity lexicons perform in a Sentiment Analysis
task. The \emph{average ratio} is used to present results of tables
\ref{tab:evalbespalov} and \ref{tab:evalmpqa} (with Bespalov corpus),
and \ref{tab:evalspanishsynsetfull}  and
\ref{tab:evalspanishfull} (with HOpinion), whereas Mohammad
\textit{et al.'s} is used to report results in table
\ref{tab:evalmpqaphrase}. Mohammad \textit{et al.'s}
\shortcite{mohammad_generating_2009} testset based on MPQA 1.1  is
smaller, but both MPQA 1.1 and 1.2 are hugely skewed towards negative polarity (30\%
positive vs. 70\% negative).

All datasets were POS tagged and Word Sense Disambiguated using FreeLing
\cite{freeling3_padro12,AgirreSoroa:2009}. Having word sense
annotated datasets gives us the opportunity to evaluate the lexicons
both at word and sense levels. For the evaluation of those lexicons
that are synset-based, such as \emph{qwn-ppv} and SentiWordNet 3.0, we
convert them from senses to words by taking every word or variant
contained in each of their senses. Moreover, if a lemma appears as a
variant in several synsets the most frequent polarity is assigned to
that lemma.

With respect to lexicons at word level, we take the most
frequent sense according to WordNet 3.0 for each of their positive and
negative words. Note that the latter
conversion, for synset based evaluation, is mostly done to show
that the evaluation at synset level is harder independently of the
quality of the lexicon evaluated. 

\subsection{Results}\label{sec:results}

Although tables \ref{tab:evalbespalov}, \ref{tab:evalmpqa} and
\ref{tab:evalmpqaphrase} also present results at synset level, it
should be noted that the only polarity lexicons available to us for
comparison at synset level were Q-WordNet \cite{agerri_q-wordnet:_2010} and
SentiWordNet 3.0 \cite{baccianella_sentiwordnet_2010}. \emph{QWN-PPV-AG} refers
to the lexicon generated starting from \textbf{AG}'s seeds, 
and \emph{QWN-PPV-TL} using \textbf{TL}'s seeds as described in section
\ref{sec:qwn:-seed-generation}. Henceforth, we will use \emph{qwn-ppv} to refer to
the overall method presented in this paper, regardless of the seeds
used.

For every \emph{qwn-ppv} result reported in this section, we have used every graph described in
section \ref{sec:ppv-generation}. The configuration of each \emph{qwn-ppv} in
the results specifies which seed iteration is used as the initialization 
of the Personalized PageRank algorithm, and on which graph.  
Thus, QWN-PPV-TL (s05\_G4) in table \ref{tab:evalmpqa} means that
the 5th iteration of synset seeds was used to propagate over graph G4. 
If the configuration were (w05\_G4) it would have meant `the 5th iteration of word
seeds were used to propagate over graph G4'. The simplicity of our approach
allows us to generate many lexicons simply by projecting a LKB over
different graphs. 

The lexicons marked with an asterisk denote those that have been converted from word to senses using the
most frequent sense of WordNet 3.0. We would like
to stress again that the purpose of such word to synset conversion is
to show that SA tasks at synset level are harder than at word level. In
addition, it should also be noted that in the case of SO-CAL
\cite{taboada_lexicon-based_2010}, we have reduced what is a graded
lexicon with scores ranging from 5 to -5 into a binary one.

Table \ref{tab:evalbespalov} shows that (at least partially)
manually built lexicons obtain the best results on this
evaluation. It also shows that \emph{qwn-ppv}
clearly outperforms any other automatically built lexicons. Moreover,
manually built lexicons suffer from the evaluation at synset level,
obtaining most of them lower scores than \emph{qwn-ppv}, although Liu's
\cite{hu_mining_2004} still obtains the best results. In any case, for
an unsupervised procedure, \emph{qwn-ppv} lexicons obtain
very competitive results with respect to manually created lexicons and
is the best among the automatic methods. It should also be noted that
the best results of \emph{qwn-ppv} are obtained with graph G1 and with
very few seed iterations. 

Table \ref{tab:evalmpqa} again sees the manually built lexicons
performing better although overall the differences are lower with
respect to automatically built lexicons. Among these, \emph{qwn-ppv}
again obtains the best results, both at synset and word level, although
in the latter the differences with MSOL(ASL-GI) are not large. Finally, table
\ref{tab:evalmpqaphrase} shows that \emph{qwn-ppv} again outperforms
other automatic approaches and is closer to those have been (partially
at least) manually built. In both MPQA evaluations the best graph
overall to propagate the seeds is G3 because this type of task favours
high recall. 

\begin{table}[ht]\footnotesize
\centering
\begin{tabular}{p{2cm}p{0.6cm}p{0.2cm}p{0.2cm}p{0.2cm}p{0.2cm}p{0.2cm}p{0.2cm}} \hline
  & & \multicolumn{3}{c}{Positives} & \multicolumn{3}{c}{Negatives} \\
\textbf{Lexicon} & size & P & R & F & P & R & F \\ \hline \hline
\multicolumn{8}{l}{\scriptsize{\textit{Automatically created}}} \\
SWN & 27854 & .87 & .99 & .93 & .70 & .16 & .27\\
QWN-PPV-AG (wrd01\_G1) & 3306 & .86 & .00 & .92 & .67 & .01 & .02 \\
QWN-PPV-TL (s04\_G1)& 5010 & .89 & .96 & {\bf.93} & .58 & .30 & {\bf.39} \\ \hline
\end{tabular}
\caption{Evaluation of Spanish lexicons using the full HOpinion corpus at synset level.}\label{tab:evalspanishsynsetfull}
\end{table}

We report results on the Spanish HOpinion corpus in tables
\ref{tab:evalspanishsynsetfull} and \ref{tab:evalspanishfull}. Mihalcea(f) is
a manually revised lexicon based on the automatically built
Mihalcea(m) \cite{perez2012learning}. ElhPolar \cite{ElhPolar_Saralegi2013} is
semi-automatically built and manually corrected. SO-CAL is built
manually. SWN and QWN-PPV have been built via the MCR 3.0's ILI 
by applying the synset to word conversion previously
described on the Spanish dictionary of the MCR. The results for Spanish
at word level in table \ref{tab:evalspanishfull} show 
the same trend as for English: \emph{qwn-ppv} is the best of the
automatic approaches and it obtains competitive although not as good as the best 
of the manually created lexicons (ElhPolar). Due to the disproportionate
number of positive reviews, the results for the negative polarity are
not useful to draw any meaningful conclusions. Thus, we also performed an evaluation
with HOpinion Balanced set as listed in table \ref{tab:datasets}. 

\begin{table}[ht]\footnotesize
\centering
\begin{tabular}{p{2cm}p{0.5cm}p{0.3cm}p{0.3cm}p{0.3cm}p{0.3cm}p{0.3cm}p{0.3cm}} \hline
 & & \multicolumn{3}{c}{Positives} & \multicolumn{3}{c}{Negatives} \\
\textbf{Lexicon} & size & P & R & F & P & R & F \\ \hline \hline
\multicolumn{8}{l}{\scriptsize{\textit{Automatically created}}} \\
Mihalcea(m) & 2496 & .86 & .00 & .92 & .00 & .00 & .00\\
SWN & 9712 & .88 & .97 & .92 & .55 & .19 & .28\\
QWN-PPV-AG (s11\_G1) & 1926 & .89 & .97 & .93 & .59 & .26 & .36\\ 
QWN-PPV-TL (s03\_G1) & 939 & .89 & .98 & {\bf.93} & .71 & .26 & {\bf.38}\\ \hline\hline
\multicolumn{8}{l}{\scriptsize{\textit{(Semi-) Manually created}}}\\
ElhPolar & 4673 & .94 & .94 & {\bf .94} & .64 & .64 & {\bf.64}\\
Mihalcea(f) & 1347 & .91 & .96 & .93 & .61 & .41 & .49\\
SO-CAL & 4664 & .92 & .96 & .94 & .70 & .51 & .59\\ \hline
\end{tabular}
\caption{Evaluation of Spanish lexicons using the full HOpinion corpus
  at word level.}\label{tab:evalspanishfull}
\end{table}

The results with a balanced HOpinion, not shown due to lack of space, 
also confirm the previous trend: \emph{qwn-ppv} outperforms other automatic
approaches but is still worse than the best of the
manually created ones (ElhPolar). 

\subsection{Intrinsic evaluation}\label{sect:intrinsic-eval}

To facilitate intrinsic comparison with previous approaches, 
we evaluate our automatically generated lexicons against GI. For each
\emph{qwn-ppv} lexicon shown in previous extrinsic evaluations, we compute the intersection
between the lexicon and GI, and evaluate the words in that
intersection. Table \ref{tab:intrinsic-eval} shows results for the
best-performing QWN-PPV lexicons (both using AG and TL seeds) 
in the extrinsic evaluations at word level of tables
\ref{tab:evalbespalov} (first two rows), \ref{tab:evalmpqa} (rows 3 and
4) and \ref{tab:evalmpqaphrase} (rows 5 and 6). 
We can see that QWN-PPV lexicons systematically outperform SWN in number
of correct entries. {\it QWN-PPV-TL} lexicons obtain 75.04\% of correctness on
average. The best performing lexicon contains up to 81.07\% of correct
entries. Note that we did not compare the results with MSOL(ASL-GI)
because it contains the GI.   

\begin{table}[ht]\footnotesize
\centering
\begin{tabular}{lcp{0.3cm}p{0.2cm}p{0.4cm}} \hline
\textbf{Lexicon} & {\bf $\cap$ wrt. GI} & {\bf Acc.} & {\bf Pos} & {\bf Neg } \\ \hline \hline
SWN & 2,755 & .74 & .76 & .73 \\
QWN-PPV-AG (w01\_G1) & 849 & .71 & .68 & .75 \\ 
QWN-PPV-TL (w01\_G1) & 713 & .78 & .80 & .76 \\ 
QWN-PPV-AG (s09\_G4) & 3,328 & .75 & .75 & .77 \\ 
QWN-PPV-TL (s05\_G4) & 3,333 & \textbf{.80} & \textbf{.84} & \textbf{.77} \\ 
QWN-PPV-AG (w02\_G3) & 3,340 & .74 & .71 & .77 \\ 
QWN-PPV-TL (s06\_G3) & 3,340 & .77 & .79 & .77 \\ \hline
\end{tabular}
\caption{Accuracy QWN-PPV lexicons and SWN with
  respect to the GI lexicon.}\label{tab:intrinsic-eval}
\end{table}

\subsection{Discussion}\label{sec:discussion}

\emph{QWN-PPV} lexicons obtain the best results among the
evaluations for English and Spanish. Furthermore, across tasks and datasets \emph{qwn-ppv}
provides a more consistent and robust behaviour than most of the manually-built
lexicons apart from OF. The results also show that for a task requiring
high recall the larger graphs, e.g. G3, are preferable, whereas for a
more balanced dataset and document level task smaller G1 graphs perform better. 

These are good results considering that our method to generate
\emph{qwn-ppv} is simpler, more robust and adaptable than
previous automatic approaches. Furthermore, although also based on a Personalized PageRank
application, it is much simpler than SentiWordNet 3.0,  
consistently outperformed by \emph{qwn-ppv} on every evaluation and
dataset. The main differences with respect to
SentiWordNet's approach are the following: (i) the seed generation and
training of 7 supervised classifiers corresponds in \emph{qwn-ppv} to
only one simple step, namely, the automatic
generation of seeds as explained in section \ref{sec:qwn:-seed-generation}; (ii) the
generation of \emph{qwn-ppv} only requires a LKB's graph for the
Personalized PageRank propagation, no disambiguated glosses; (iii) the
graph they use to do the propagation also depends on disambiguated
glosses, not readily available for any language.  

The fact that \emph{qwn-ppv} is based on already available WordNets
projected onto simple graphs is crucial for the robustness and adaptability of the \emph{qwn-ppv}
method across evaluation tasks and datasets: Our method can quickly create, over
different graphs, many lexicons of diffent sizes which can
then be evaluated on a particular polarity classification task and dataset. 
Hence the different configurations of the \emph{qwn-ppv} lexicons, 
because for some tasks a G3 graph with more AG/TL seed iterations will obtain better recall and
viceversa. This is confirmed by the results: the tasks using MPQA seem to
clearly benefit from high recall whereas the Bespalov's corpus has overall, 
more balanced scores. This could also be due to the size of Bespalov's corpus, 
almost 10 times larger than MPQA 1.2. 

The experiments to generate Spanish lexicons confirm the trend showed
by the English evaluations: Lexicons generated by \emph{qwn-ppv}
consistenly outperform other automatic approaches, although some manual
lexicon is better on a given task and dataset (usually a different one). Nonetheless
the Spanish evaluation shows that our method is also robust across
languages as it gets quite close to the manually corrected lexicon of
Mihalcea(full) \cite{perez2012learning}.

The results also confirm that no single lexicon is the most
appropriate for any SA task or dataset and domain. In this sense,
the adaptability of \emph{qwn-ppv} is a desirable feature for lexicons
to be employed in SA tasks: the unsupervised \emph{qwn-ppv} method only relies on
the availability of a LKB to build hundreds of polarity lexicons which can then
be evaluated on a given task and dataset to choose the best fit. If not
annotated evaluation set is available, G3-based propagations provide the
best recall whereas the G1-based lexicons are less noisy. Finally, we believe that the results
reported here point out to the fact that intrinsic evaluations are not meaningful to judge the
adequacy a polarity lexicon for a specific SA task.

\section{Concluding Remarks}\label{sec:concluding-remarks}

This paper presents an unsupervised dictionary-based method \emph{qwn-ppv} to
automatically generate polarity lexicons. Although simpler than
similar automatic approaches, it still obtains better
results on the four extrinsic evaluations presented. Because it only depends on the
availability of a LKB, we believe that this method can be
valuable to generate on-demand polarity lexicons for a given language
when not sufficient annotated data is available. We demonstrate the adaptability of
our approach by producing good performance polarity lexicons for
different evaluation scenarios and for more than one language. 

Further work includes investigating different graph projections of
WordNet relations to do the propagation as well as exploiting
synset weights. We also plan to investigate the use of
annotated corpora to generate lexicons at word level to try and close
the gap with those that have been (at least partially) manually
annotated.

The \emph{qwn-ppv} lexicons and graphs
used in this paper are publicly available (under CC-BY license):
\emph{http://www.rodrigoagerri.net/resources/\#qwn-ppv}. The \emph{qwn-ppv tool} to automatically
generate polarity lexicons given a WordNet in any language will soon be
available in the aforementioned URL.

\section*{Acknowledgements}

This work has been supported by the OpeNER FP7 project
under Grant No. 296451, the FP7 NewsReader project, Grant No. 316404 and by the Spanish MICINN project SKATER under Grant No. TIN2012-38584-C06-01.

\bibliographystyle{acl}

\begin{thebibliography}{}

\bibitem[\protect\citename{Agerri and
  Garc\'ia-Serrano}2010]{agerri_q-wordnet:_2010}
R.~Agerri and A.~Garc\'ia-Serrano.
\newblock 2010.
\newblock Q-{WordNet:} extracting polarity from {WordNet} senses.
\newblock In {\em Seventh Conference on International Language Resources and
  Evaluation, Malta. Retrieved May}, volume~25, page 2010.

\bibitem[\protect\citename{Agirre and Soroa}2009]{AgirreSoroa:2009}
Eneko Agirre and Aitor Soroa.
\newblock 2009.
\newblock Personalizing pagerank for word sense disambiguation.
\newblock In {\em Proceedings of the 12th Conference of the European Chapter of
  the Association for Computational Linguistics (EACL-2009)}, Athens, Greece.

\bibitem[\protect\citename{Agirre \bgroup et al.\egroup
  }2012]{agirre2012multilingual}
Aitor~Gonz{\'a}lez Agirre, Egoitz Laparra, German Rigau, and Basque~Country
  Donostia.
\newblock 2012.
\newblock Multilingual central repository version 3.0: upgrading a very large
  lexical knowledge base.
\newblock In {\em GWC 2012 6th International Global Wordnet Conference}, page
  118.

\bibitem[\protect\citename{Agirre \bgroup et al.\egroup
  }2014]{agirre2014random}
Eneko Agirre, Oier~Lopez de~Lacalle, and Aitor Soroa.
\newblock 2014.
\newblock Random walks for knowledge-based word sense disambiguation.
\newblock {\em Computational Linguistics}, (Early Access).

\bibitem[\protect\citename{Baccianella \bgroup et al.\egroup
  }2010]{baccianella_sentiwordnet_2010}
S.~Baccianella, A.~Esuli, and F.~Sebastiani.
\newblock 2010.
\newblock {SentiWordNet} 3.0: An enhanced lexical resource for sentiment
  analysis and opinion mining.
\newblock In {\em Seventh conference on International Language Resources and
  Evaluation (LREC-2010), Malta.}, volume~25.

\bibitem[\protect\citename{Bespalov \bgroup et al.\egroup
  }2011]{bespalov_sentiment_2011}
Dmitriy Bespalov, Bing Bai, Yanjun Qi, and Ali Shokoufandeh.
\newblock 2011.
\newblock Sentiment classification based on supervised latent n-gram analysis.
\newblock In {\em Proceedings of the 20th {ACM} international conference on
  Information and knowledge management}, pages 375--382.

\bibitem[\protect\citename{Bond and Foster}2013]{bond_linking_2013}
Francis Bond and Ryan Foster.
\newblock 2013.
\newblock Linking and extending an open multilingual wordnet.
\newblock In {\em 51st Annual Meeting of the Association for Computational
  Linguistics: {ACL-2013}}.

\bibitem[\protect\citename{Brin and Page}1998]{brin_anatomy_1998}
Sergey Brin and Lawrence Page.
\newblock 1998.
\newblock The anatomy of a large-scale hypertextual web search engine.
\newblock {\em Computer networks and {ISDN} systems}, 30(1):107–117.

\bibitem[\protect\citename{Choi and Cardie}2009]{choi_adapting_2009}
Y.~Choi and C.~Cardie.
\newblock 2009.
\newblock Adapting a polarity lexicon using integer linear programming for
  domain-specific sentiment classification.
\newblock In {\em Proceedings of the 2009 Conference on Empirical Methods in
  Natural Language Processing: Volume 2-Volume 2}, pages 590--598.

\bibitem[\protect\citename{Ding \bgroup et al.\egroup
  }2008]{ding_holistic_2008}
X.~Ding, B.~Liu, and P.~S. Yu.
\newblock 2008.
\newblock A holistic lexicon-based approach to opinion mining.
\newblock In {\em Proceedings of the international conference on Web search and
  web data mining}, pages 231--240.

\bibitem[\protect\citename{Esuli and Sebastiani}2007]{EsuliSebastiani:2007}
Andrea Esuli and Fabrizio Sebastiani.
\newblock 2007.
\newblock Pageranking wordnet synsets: An application to opinion mining.
\newblock In {\em Proceedings of the 45th Annual Meeting of the Association of
  Computational Linguistics}, pages 424--431, Prague, Czech Republic, June.
  Association for Computational Linguistics.

\bibitem[\protect\citename{Fellbaum and Miller}1998]{Wordnet:1998}
C.~Fellbaum and G.~Miller, editors.
\newblock 1998.
\newblock {\em Wordnet: An Electronic Lexical Database}.
\newblock MIT Press, Cambridge (MA).

\bibitem[\protect\citename{Hatzivassiloglou and
  {McKeown}}1997]{hatzivassiloglou_predicting_1997}
V.~Hatzivassiloglou and K.~R {McKeown}.
\newblock 1997.
\newblock Predicting the semantic orientation of adjectives.
\newblock In {\em Proceedings of the eighth conference on European chapter of
  the Association for Computational Linguistics}, pages 174--181.

\bibitem[\protect\citename{Hu and Liu}2004]{hu_mining_2004}
M.~Hu and B.~Liu.
\newblock 2004.
\newblock Mining and summarizing customer reviews.
\newblock In {\em Proceedings of the tenth {ACM} {SIGKDD} international
  conference on Knowledge discovery and data mining}, pages 168--177.

\bibitem[\protect\citename{Kim and Hovy}2004]{KimHovy:2004}
Soo-Min Kim and Eduard Hovy.
\newblock 2004.
\newblock Determining the sentiment of opinions.
\newblock In {\em Proceedings of Coling 2004}, pages 1367--1373, Geneva,
  Switzerland, Aug 23--Aug 27. COLING.

\bibitem[\protect\citename{Liu}2012]{liu_sentiment_2012}
Bing Liu.
\newblock 2012.
\newblock Sentiment analysis and opinion mining.
\newblock {\em Synthesis Lectures on Human Language Technologies}, 5(1):1--167.

\bibitem[\protect\citename{Mihalcea \bgroup et al.\egroup
  }2007]{mihalcea_learning_2007}
R.~Mihalcea, C.~Banea, and J.~Wiebe.
\newblock 2007.
\newblock Learning multilingual subjective language via cross-lingual
  projections.
\newblock In {\em {Annual Meeting of the Association for Computational
  Linguistics}}, volume~45, page 976.

\bibitem[\protect\citename{Mohammad \bgroup et al.\egroup
  }2009]{mohammad_generating_2009}
S.~Mohammad, C.~Dunne, and B.~Dorr.
\newblock 2009.
\newblock Generating high-coverage semantic orientation lexicons from overtly
  marked words and a thesaurus.
\newblock In {\em Proceedings of the 2009 Conference on Empirical Methods in
  Natural Language Processing: Volume 2-Volume 2}, pages 599--608.

\bibitem[\protect\citename{Padr\'o and Stanilovsky}2012]{freeling3_padro12}
Llu\'is Padr\'o and Evgeny Stanilovsky.
\newblock 2012.
\newblock Freeling 3.0: Towards wider multilinguality.
\newblock In {\em Proceedings of the Language Resources and Evaluation
  Conference (LREC 2012)}, Istanbul, Turkey, May. ELRA.

\bibitem[\protect\citename{Pang and Lee}2008]{pang_opinion_2008}
B.~Pang and L.~Lee.
\newblock 2008.
\newblock Opinion mining and sentiment analysis.
\newblock {\em Foundations and Trends in Information Retrieval}, 2(1-2):1--135.

\bibitem[\protect\citename{P{\'e}rez-Rosas \bgroup et al.\egroup
  }2012]{perez2012learning}
Ver{\'o}nica P{\'e}rez-Rosas, Carmen Banea, and Rada Mihalcea.
\newblock 2012.
\newblock Learning sentiment lexicons in spanish.
\newblock In {\em LREC}, pages 3077--3081.

\bibitem[\protect\citename{Rao and Ravichandran}2009]{rao_semi-supervised_2009}
D.~Rao and D.~Ravichandran.
\newblock 2009.
\newblock Semi-supervised polarity lexicon induction.
\newblock In {\em Proceedings of the 12th Conference of the European Chapter of
  the Association for Computational Linguistics}, pages 675--682.

\bibitem[\protect\citename{Riloff and Wiebe}2003]{RiloffWiebe:2003}
E.~Riloff and J.~Wiebe.
\newblock 2003.
\newblock Learning extraction patterns for subjective expressions.
\newblock In {\em Proceedings of the International Conference on Empirical
  Methods in Natural Language Processing (EMNLP'03)}.

\bibitem[\protect\citename{Saralegi and
  San~Vicente}2013]{ElhPolar_Saralegi2013}
Xabier Saralegi and I\~naki San~Vicente.
\newblock 2013.
\newblock Elhuyar at {TASS}2013.
\newblock In {\em XXIX Congreso de la Sociedad Española de Procesamiento de
  lenguaje natural, Workshop on Sentiment Analysis at SEPLN (TASS2013)}, pages
  143--150, Madrid.

\bibitem[\protect\citename{Stone \bgroup et al.\egroup }1966]{Stoneetal:1966}
P.~Stone, D.~Dunphy, M.~Smith, and D.~Ogilvie.
\newblock 1966.
\newblock {\em The General Inquirer: A Computer Approach to Content Analysis}.
\newblock Cambridge (MA): MIT Press.

\bibitem[\protect\citename{Strapparava and
  Valitutti}2004]{StrapparavaValitutti:2004}
Carlo Strapparava and Alessandro Valitutti.
\newblock 2004.
\newblock Wordnet-affect: an affective extension of wordnet.
\newblock In {\em Proceedings of the 4th International Conference on Languages
  Resources and Evaluation (LREC 2004)}, pages 1083--1086, Lisbon, May.

\bibitem[\protect\citename{Taboada \bgroup et al.\egroup
  }2010]{taboada_lexicon-based_2010}
M.~Taboada, J.~Brooke, M.~Tofiloski, K.~Voll, and M.~Stede.
\newblock 2010.
\newblock Lexicon-based methods for sentiment analysis.
\newblock {\em Computational Linguistics}, (Early Access):1–41.

\bibitem[\protect\citename{Takamura \bgroup et al.\egroup
  }2005]{takamura_extracting_2005}
Hiroya Takamura, Takashi Inui, and Manabu Okumura.
\newblock 2005.
\newblock Extracting semantic orientations of words using spin model.
\newblock In {\em Proceedings of the 43rd Annual Meeting of the Association for
  Computational Linguistics ({ACL'05)}}, page 133–140, Ann Arbor, Michigan,
  June. Association for Computational Linguistics.

\bibitem[\protect\citename{Turney and Littman}2003]{turney_measuring_2003}
P.~Turney and M.~Littman.
\newblock 2003.
\newblock Measuring praise and criticism: Inference of semantic oreintation
  from association.
\newblock {\em {ACM} Transaction on Information Systems}, 21(4):315--346.

\bibitem[\protect\citename{Turney}2002]{turney_thumbs_2002}
{P.D.} Turney.
\newblock 2002.
\newblock Thumbs up or thumbs down?: semantic orientation applied to
  unsupervised classification of reviews.
\newblock In {\em Proceedings of the 40th Annual Meeting on Association for
  Computational Linguistics}, page 417–424.

\bibitem[\protect\citename{Wilson \bgroup et al.\egroup
  }2005]{wilson_recognizing_2005}
Theresa Wilson, Janyce Wiebe, and Paul Hoffmann.
\newblock 2005.
\newblock Recognizing contextual polarity in phrase-level sentiment analysis.
\newblock In {\em Proceedings of the conference on Human Language Technology
  and Empirical Methods in Natural Language Processing}, page 347–354.

\end{thebibliography}

\end{document}